\documentclass{article}

\usepackage{PRIMEarxiv}

\usepackage[utf8]{inputenc} % allow utf-8 input
\usepackage[T1]{fontenc}    % use 8-bit T1 fonts
\usepackage{hyperref}       % hyperlinks
\usepackage{url}            % simple URL typesetting
\usepackage{booktabs}       % professional-quality tables
\usepackage{amsfonts}       % blackboard math symbols
\usepackage{nicefrac}       % compact symbols for 1/2, etc.
\usepackage{microtype}      % microtypography
\usepackage{lipsum}
\usepackage{fancyhdr}       % header
\usepackage{graphicx}       % graphics
\graphicspath{{media/}}     % organize your images and other figures under media/ folder

\usepackage{amsmath}
\usepackage{appendix}

\newtheorem{theorem}{Theorem}

\title{Recurrent Joint Embedding Predictive Architecture with Recurrent Forward Propagation Learning
}

\author{
  Osvaldo M Velarde\\
  Biomedical Engineering Department \\
  The City College of New York \\
  \texttt{ovelarde@ccny.cuny.edu} \\
  \And
    Lucas C Parra \\
  Biomedical Engineering Department \\
  The City College of New York \\
  \texttt{parra@ccny.cuny.edu} \\
}

\begin{document}
\maketitle

\begin{abstract}
    Conventional computer vision models rely on very deep, feedforward networks processing whole images and trained offline with extensive labeled data. In contrast, biological vision relies on comparatively shallow, recurrent networks that analyze sequences of fixated image patches, learning continuously in real-time without explicit supervision. This work introduces a vision network inspired by these biological principles.  Specifically, it leverages a joint embedding predictive architecture \cite{garrido_learning_2024} incorporating recurrent gated circuits \cite{nayebi_recurrent_2022}.  The network learns by predicting the representation of the next image patch (fixation) based on the sequence of past fixations, a form of self-supervised learning. We show mathematical and empirically that the training algorithm avoids the problem of representational collapse. We also introduce \emph{Recurrent-Forward Propagation}, a learning algorithm that avoids biologically unrealistic backpropagation through time \cite{werbos_backpropagation_1990} or memory-inefficient real-time recurrent learning \cite{williams_learning_1989}. We show mathematically that the algorithm implements exact gradient descent for a large class of recurrent architectures, and confirm empirically that it learns efficiently. This paper focuses on these theoretical innovations and leaves empirical evaluation of performance in downstream tasks, and analysis of representational similarity with biological vision for future work.  
\end{abstract}

% keywords can be removed

\section{Introduction}
One of the most significant challenges for our visual system is the need to integrate visual information across the constant movements of our eyes \cite{kong_transsaccadic_2021,oostwoud_wijdenes_evidence_2015}. We move our eyes in rapid saccades several times per second, holding fixation only for fractions of a second, yet our brain constructs a coherent and stable representation of the visual scene. How the brain accomplishes this remains an open question. 

Visual processing in the primate brain is structured in consecutive visual processing areas that extract increasingly more complex features of the visual input \cite{vision_information_1990}. In every area, the visual field is processed similarly across locations. This structure has been emulated with artificial neural networks, organized in consecutive layers, with convolutions at every layer emulating the spatially invariant processing \cite{celeghin_convolutional_2023}. A deep stack of such convolutional neural networks (CNNs) achieves impressive results in static image analysis tasks; however, they do not take into account the temporal dynamics inherent in visual processing. For processing sequential data, Recurrent Neural Networks (RNNs) are one of the preferred models due to their ability to model long-range dependencies and capture the evolution of temporal patterns over time \cite{schmidt_recurrent_2019}. Indeed, biological vision is dominated by recurrent feedback, both within brain area as well as between areas \cite{gilbert_top-down_2013}. Therefore, the combination of recurrent connections and convolutional layers could effectively address challenges such as video understanding, action recognition, and object tracking. There are a few recent efforts to incorporate recurrence into vision models \cite{nayebi_recurrent_2022,bai_empirical_2018,alom_inception_2017,donahue_long-term_2016}. However, it is not clear how such networks should handle visual input that is constantly changing due to eye movements, or how such networks should be trained. 

The conventional approach to training computer vision models is supervised learning, i.e., using images as input and labels for those images as output. However, humans and animals do not receive explicit labels for every object in a scene; instead, they learn patterns, relationships, and regularities from continuous experience. Most learning occurs implicitly, which is more consistent with Self-Supervised Learning (SSL) techniques. SSL is a paradigm in machine learning where a model uses the input itself as targets for learning. Often, the model is trained to predict an unobserved part of the input from the observed parts. In this way, the model can learn to relate multiple "views" or multiple modalities in large datasets without relying on explicit labels (e.g., predict images from associated audio or text).

In this work, we present a new biologically inspired architecture that integrates convolutional layers, recurrent networks, and SSL using next-step prediction. Specifically, the network is trained to predict a representation of a fixated image patch, based on previous fixations. In doing so, the recurrent architecture can and should integrate information across multiple fixations, thus forming an wholistic representation of a scene. Our model follows the joint-embedding predictive architecture (JEPA) \cite{assran_self-supervised_2023,dawid_introduction_2024} and extends this to include recurrent networks, in short, R-JEPA. 

The specific network architecture we propose here borrows from previous efforts to train vision networks with contrastive \cite{chen_simple_2020} or predictive SSL \cite{chen_exploring_2020}, essentially, a ResNet50. The recurrent structure we use borrows from previous recurrent vision models that have used supervised learning \cite{nayebi_recurrent_2022} or contrastive learning \cite{zhuang_unsupervised_2021}, which still requires labels in the form of same/different exemplars. We omit the need for labels entirely, by using next-step prediction as a SSL objective.

The two main contributions of this paper are theoretical. First, we show mathematically, and confirm in simulations that the R-JEPA avoids the problem of representational collapse, despite lacking negative exemplars used for contrastive learning or otherwise enforcing diversity of representations, as in \cite{zbontar_barlow_2021, bardes_vicreg_2022}. 
Second, we will show that the network can be trained in real-time with a forward propagation of sensitivity for each weight in the recurrent network, rather than costly back-propagation through time, usually required for recurrent networks. We show the conditions under which this efficient \textit{Recurrent Forward Propagation} is an exact gradient computation. Without realizing these conditions, previous work on efficient real-time recurrent learning only offered approximate solutions \cite{marschall_unified_2019,irie_exploring_2024,gerstner_eligibility_2018,bellec_solution_2020}. We confirm in simulation that the network learns effectively when applying Recurrent Forward Propagation in real-time. 

\section{Methods}
\subsection{Recurrent Joint Embedding Predictive Architecture}
\begin{figure}[ht]
\centering
\includegraphics[scale=0.25]{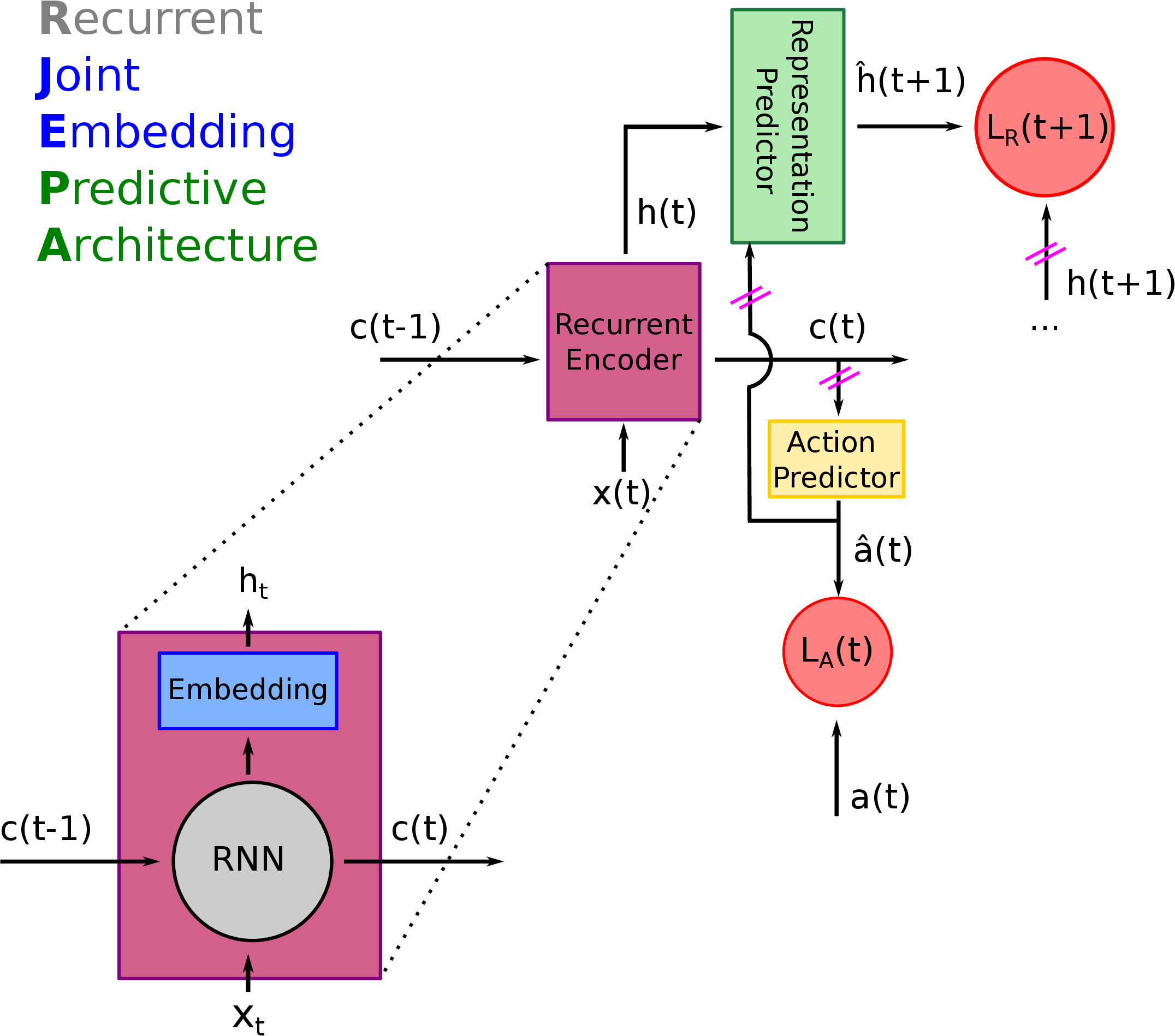}
\caption{\textbf{Recurrent Joint Embedding Predictive Architecture (R-JEPA)}. For an input $x(t)$, the encoder generates a representation $h(t)$. Then, the predictor generates a prediction $\hat{h}(t+1)$ of the representation of the next input. The objective of the encoder and predictor is to minimize the prediction loss $\mathcal{L}_R$ in the embedding space $\mathcal{H}$. At the same time, the context vector $c(t)$ is used to predict the action/response $\hat{a}(t)$ to stimuli $x(t)$.}
\label{fig:summary}
\end{figure}

R-JEPA is a novel self-supervised recurrent model for the processing of high-dimensional time series $x(t)$ (see Figure \ref{fig:summary}). Like JEPA \cite{assran_self-supervised_2023}, our model has two key elements:

\begin{enumerate}
    \item \textbf{embedding}: The architecture defines an embedding space $\mathcal{H}$ through  function $\text{Enc}:\mathcal{X} \rightarrow \mathcal{H}$ that transform the trajectory $x(t)$ of a high-dimensional space $\mathcal{X}$ to a trajectory $h(t)$ in $\mathcal{H}$.
    
    \item \textbf{predictive architecture}: It can learn to predict $h(t)$ from $h(t-\Delta)$ with a function $G:\mathcal{H} \rightarrow \mathcal{H}$.
\end{enumerate}

The representations in the embedding space $h(t)$ must satisfy the following properties: (1) they should be maximally informative about $x(t)$, and (2) $h(t)$ must be easily predictable from $h(t-\Delta)$. The point (1) prevents \textit{representational collapse} (i.e. $h(t)$ is a point in $\mathcal{H}$, or weakly informative). On the other hand, point (2) is enforced by a loss function $\mathcal{L}_R$ defined in the embedding space $\mathcal{H}$ to capture the quality of the prediction. 

Our goal for R-JEPA is to mimic and extend how humans perceive, and comprehend the world around them, and anticipate future events based on experiences. We expect R-JEPA to be able to learn temporal dynamics and causal relationships. Here we develop the concept in the context of vision, but JEPA has more generally been used in a multimodal context of vision, hearing and action. 

R-JEPA is trained to predict representations of the future from representations of the past and present. For this, we use an encoder \textit{Enc} based on Recurrent Neural Networks (RNNs). RNNs compute time-variant internal states whose transition is determined by the information from the past and input at present. The context vector $c(t)$ represents accumulated past information that the RNN uses for processing the new input $x(t)$. We select a context vector that is composed of two components $c(t) = (s(t),m(t))$, where $s(t)$ is the internal state of the network and $m(t)$ is a memory signal. Finally, the trajectory $h(t)$ in the embedding space is a projection of the context vector $c(t)$.

Mathematically,

\begin{align*}
c(t) &=  C(x(t), c(t-1); \theta^{(C)}) & \text{Enconder: RNN} \\
h(t) &=  F(c(t); \theta^{(F)}) & \text{Enconder: Embedding} \\
\hat{a}(t) &=  A(c(t); \theta^{(A)}) & \text{Action Predictor} \\
\hat{h}(t+1) &=  G(h(t),\hat{a}(t);\theta^{(G)}) & \text{Representation Predictor}\\
\mathcal{L}_R(t) &= d_R(h(t),\hat{h}(t)) & \text{Loss of representations} \\
\mathcal{L}_A(t) &= d_A(a(t),\hat{a}(t)) & \text{Loss of actions} \\
E &= \mathbb{E}_t[\mathcal{L}_R(t) + \mathcal{L}_A(t)]
\end{align*}

Like JEPA, R-JEPA can also produce a \textit{behavioral response} to input stimuli $x(t)$, which we’ll refer to as \textit{actions}. For example, an action might be an eye movement (e.g., saccades) while viewing a static image. To infer an action $\hat{a}(t)$, information from the context vector $c(t)$ will be helpful, as this will aid in predicting the next representation $\hat{h}(t+1)$. In certain experiments, this action $a(t)$ can be measured and compared to the model's predictions $\hat{a}(t)$. For this work, we will assume that the sequence of actions is given, i.e. the eye movements are given to us, and we leave scan-path prediction or “active vision” to future work. Effectively, we assume the sequence of fixated image patches $x(t)$ is given.  

\subsection{R-JEPA is trained to predict the content of the next fixation}

\begin{figure}[ht]
\centering
\includegraphics[scale=0.20]{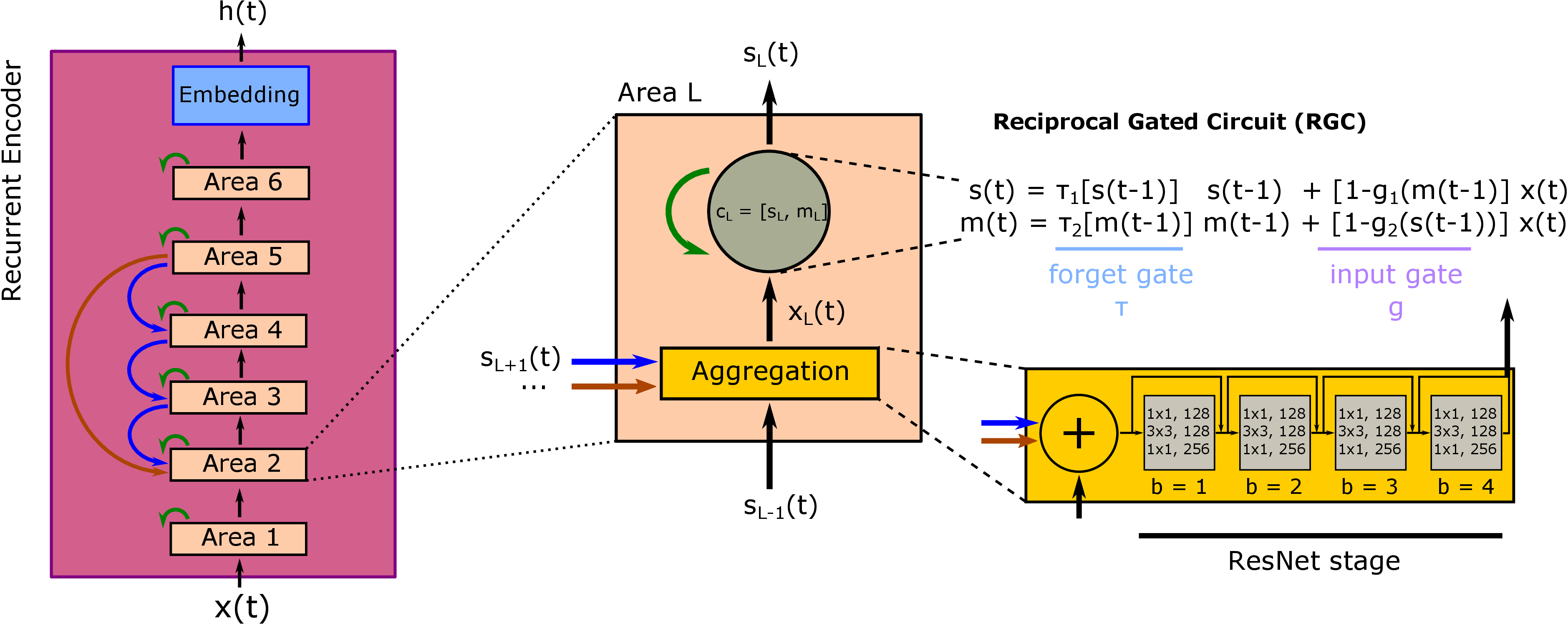}
\caption{\textbf{Recurrent Encoder}. We implemented an encoder based on ResNet50 \cite{chen_exploring_2020} and Reciprocal Gated Circuit \cite{nayebi_recurrent_2022}. The encoder is a hierarchical architecture: (1) The first five areas correspond to ResNet stages, (2) Area 6 works as a head in ResNet, and (3) Embedding is a linear projection. Areas 1-6 have recurrent units inside them.}
\label{fig:encoder}
\end{figure}

We focus on an architecture for processing the temporal sequences of fixated image patches. The encoder consists of six processing areas to allow for learning of hierarchical representations typical for models of biological vision (see Fig. \ref{fig:encoder}) \cite{yamins_using_2016,nayebi_task-driven_2018,nayebi_recurrent_2022}.

In this work, we use reciprocal gated circuits (RGCs) \cite{nayebi_recurrent_2022}:

\begin{align*}
    s(t) &= [1-\sigma(W_{ms} m(t-1))] x(t) + \sigma(W_{ss} s(t-1)) s(t-1) \\
    m(t) &= [1-\sigma(W_{sm} s(t-1))] x(t) + \sigma(W_{mm} m(t-1)) m(t-1).    
\end{align*}
where $\sigma$ is hyperbolic tangent function. 

These circuits have two properties:
\begin{enumerate}
    \item \textbf{Direct passthrough}, if the internal state is initialized to zero at the first time step $s(0)=m(0)=0$, then the RGC allows feedforward input to pass on to the next area as in a standard feedforward network: $s(1) = x(1)$. This allows the network to provide a fast initial repose, which can be refined with recurrent iteration in time. 
    
    \item \textbf{Gating}, in which the value of a internal state determines how much of the bottom-up input is passed through, retained, or discarded at the next time step. RNNs with a gating mechanism can learn long-term sequential data. 
\end{enumerate}

These properties have direct analogies to biological mechanisms: direct passthrough would correspond to feedforward processing in time, and gating would correspond to adaptation to stimulus statistics across time \cite{nayebi_recurrent_2022}.

We call the first five areas \textit{low-level areas}; while the last \textit{high-level area}. The state of the network at time $t \in \{1,2,...,T\}$ is defined by

\begin{enumerate}
    \item \textbf{Low Level:} $c^{Low}_{l}(t) = C_l(c^{Low}_{l}(t-1),x^{Low}_{l}(t))$,
    \item \textbf{High Level:} $c^{High}(t) = C_H(c^{High}(t-1),x^{High}(t))$.
\end{enumerate}

They differ in that low-level areas have spatial resolution (corresponding to retinotopy) and spatially invariant processing implemented with convolutions, whereas higher-level areas are in a feature space without explicit spatial resolution implemented with ``dense" connections. 

In both $C_l$ and $C_H$ dynamics, $x$ represents the external input. For the low-level, $x^{Low}_{l}(t)$ is the integration of information from other areas,
\begin{equation} \label{eq:ex_aggr}
    x_{l}(t) = F_{l}\left( c^{Low}_{l-1}(t-1) + \phi_l\left(\bigoplus_{k=l+1}^{N} S(c^{Low}_{k}(t-1))\right)\right).
\end{equation}

$F_{l}$ is a ``ResNet stage'' which involves downsampling, $S$ is an upsampling operation to match spatial dimensions of activity from later areas $h_{k,t}$, $\bigoplus$ indicates concatenation along the channel axis, and $\phi_l$ is a linear map combining the concatenated channels and reducing the channel dimension to match $c^{Low}_{l-1}(t)$. 

For the high-level, $x^{High}(t)$ is a compression $g$ of the collection of $c^{Low}_{l}(t)$:
\begin{equation}
x^{High}(t) = g(\{c^{Low}_{l}(t)\}) = \text{MLP}(\text{AvPool}(c^{Low}_{N}(t))).    
\end{equation}

The activity $c^{Low}_{l}(t)$ of each area $l$ is a tensor with dimensions height $\times$ width $\times$ channels. The activity $c^{High}(t)$ and the input $x^{High}(t)$ are vectors with the same dimension. The context vector $c(t)$ is the list $[c^{Low}_{1}(t), c^{Low}_{2}(t), ...,c^{Low}_{N}(t),c^{High}(t)]$.  The temporal dynamic is initialized with $c^{Low}_{l}(0) = 0$ and $c^{High}_{l}(0) = 0$, while the external input at the lowest area of the network is $c_{0}(t)=x(t)$. 

Finally, we propose embedding $F$ as a linear projection of $c_{t}$: $h_t = s^{High}_{t}$; while the representation predictor $G$ is a multilayer perceptron. We ignore the effect of the actions and action predictor in this manuscript.

\section{Result: R-JEPA avoids representational collapse}

\subsection{Balancing of the Representation Predictor and R-Enconder}

The general idea of joint embeddings is that related stimuli should have representational embeddings that are predictable from one another \cite{dawid_introduction_2024}. When learning a joint embedding, a standard problem is that of representational collapse, namely, all stimuli have the same representation. To prevent this trivial collapse, contrastive learning also aims to increase the contrast of the embedding for stimuli that are not related, e.g. patches from different images \cite{chen_simple_2020}. To avoid the need for same/different labels a number of methods attempt to generate embeddings that generated representations that are decorrelated and variance is preserved \cite{zbontar_barlow_2021,bardes_vicreg_2022}. Another approach has been to utilize architectural asymmetries (e.g., dual networks in \cite{grill_bootstrap_2020}, referred to as BYOL), or asymmetries in data modality \cite{garrido_learning_2024}. A particularly simple solutions, termed SimSiam, uses identical (Siamese) encoder networks but prevents gradient propagation in one of the encoder networks. That ``stop-gradient" alone was found to prevent representational collapse \cite{chen_exploring_2020}. Subsequent theoretical work showed this trick preserves a ``balance" between the encoder and predictor, and that this balance prevents collapse \cite{liu_bridging_2022}. Here we have implemented stop-gradient in branches of the architecture to prevent collapse (see pink lines in Fig. \ref{fig:summary}). 

In this section, we will show that for R-JEPA, a balancing of the Representation Predictor and R-Enconder exists. It ensures that the encoder will also learn what the predictor learns, which is important as the encoder's representations are what is used for downstream tasks. 

\begin{theorem}
In R-JEPA, when minimizing the square prediction error $E$ under a stable network dynamic, a linear representation predictor $W_{Gh}$ converges after repeated gradient decent iteration with weight decay to the following proportionality: 
\begin{equation}
W_{Gh}^T W_{Gh} \propto H H^T.    
\end{equation}
\end{theorem}

Here $H H^T$ measures the covariance of the representations $h(t)$ across time. The proof for this theorem for the R-JEPA is in Appendix A, and follows \cite{liu_bridging_2022}. As we discuss in the next section, in practice  matrix $W_{Gh}^T W_{Gh}$ is approximately diagonal. This implies that the representations are decorrelated after iterating gradient descent, achieving a diverse representation without explicitly requiring decorrelation, or constrained variance. While we derived this Theorem for the square prediction error, it applies equally for other error measured, such as the cosine-distance. 

\subsection{Empirical demonstration of diverse representations}

Since $W_{Gh}$ is typically initialized to random values (i.e. approximately orthogonal), the matrix $W_{Gh}^T W_{Gh}$ is approximately diagonal at the start of learning. In practice the stop-gradient in SimSiam as well as the BYOL \cite{liu_bridging_2022,zhang_how_2022} both maintain this orthogonality during gradient updates, thus maintaining a diverse representation. 

To demonstrate that R-JEPA with next-step prediction in practice also maintains a diverse representation of the input sequence we trained the network on a large dataset of fixated images. The dataset comprises sequences of image patches gathered from human participants as they watched full-length movies. The dataset includes 238 unique participant-movie combinations, with each movie lasting between 93 and 142 minutes. For each participant-movie pair, there is a specific number of recorded saccades, denoted as $n_{sac}$. Altogether, the dataset contains approximately $3.4 \times 10^6$ individual saccades. Each patch covers approximately a visual angle of 5 degrees (50 x 50  pixels). 

We trained R-JEPA to minimize the prediction error on this data. In the initial implementation, the recurrence was only at the highest area of the network (green arrow in Fig.~\ref{fig:encoder}), and the rest of the network is a feedforward ResNet50 (blue and red arrows in Fig.~\ref{fig:encoder} are omitted). It uses the same pre-trained parameters as SimSiam \cite{chen_exploring_2020}. 
The proof of the Theorem requires gradient descent with weight decay, however, in practice, weight decay is not necessary \cite{liu_bridging_2022} and we have not used it here. 

\begin{figure}[ht]
\centering
\includegraphics[scale=0.25]{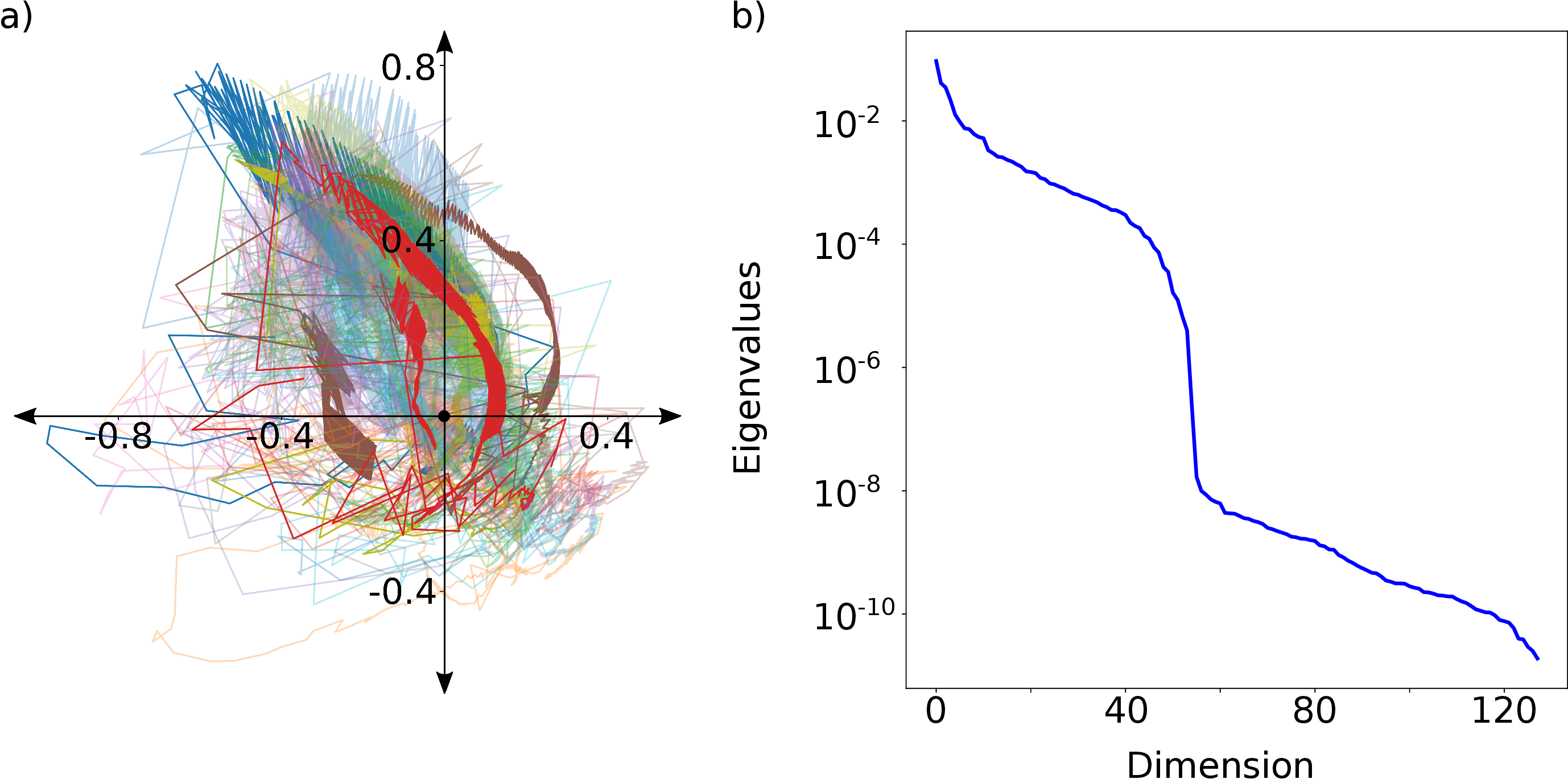}
\caption{\textbf{R-JEPA avoids collapse}. (a) A 2D projection of the trajectory of features $h(t)$ for different inputs $x(t)$ using PCA. (b) Distribution of eigenvalues of the correlation matrix of features, i.e. $H H^T$.}
\label{fig:collapse}
\end{figure}

After training the representation remains diverse  (Fig.~\ref{fig:collapse}A), filling approximately a 50 dimensional space (of the 120 dimensions of the embedding) (Fig.~\ref{fig:collapse}B). %para la prediction de la representation de 

\section{Result: R-JEPA can be trained efficiently with Forward Propagation of Sensitivity}

Biological neural networks can efficiently perform credit assignment under spatiotemporal locality constraints \cite{ellenberger_backpropagation_2024}. In machine learning, Backpropagation (BP) is the standard solution to the credit assignment problem in feedforward networks \cite{rumelhart_learning_1986} and BP-through-time (BPTT) for recurrent networks \cite{werbos_backpropagation_1990}. Both methods rely on biologically implausible assumptions \cite{lillicrap_backpropagation_2019,hebb_organization_2002}, particularly regarding temporal locality. For example, BPTT cannot be performed online, as errors are calculated retrospectively only at the end of a task (i.e. processing of the complete time sequence $\{x(t)\}^T_{t=1}$). This requires either storing the entire network history and/or recomputing it over again at every update \cite{ellenberger_backpropagation_2024}. 

There are alternative gradient forward propagation algorithms \cite{marschall_unified_2019}, which are temporally local, referred to online or real-time learning. \cite{williams_learning_1989} presents one of the canonical examples called Real-Time Recurrent Learning (RTRL) that computes how each parameter affects the hidden states at each time step. However, this tensor $\gamma$ increases with order $\mathcal{O}(n^3)$, with $n$ being the number of nodes in the network. This makes RTRL  computationally prohibitive in large networks and more costly than BPTT (order $\mathcal{O}(n^2)$).

In this Section, we show that the computational cost of RTRL for recurrent gated circuits (RGCs) is only $\mathcal{O}(n^2)$. We refer to the resulting algorithms as \emph{Recurrent Forward Propagation} {\bf (RFP)}. We proof that RFP is applicable more generally to recurrent networks that contain only \textit{two-points interactions}. We then show empirical results with both BPTT and RFP.   

\subsection{Reducing the computational cost anpad memory storage of the Recurrent Gated Circuit}

Given the loss function $E = \mathbb{E}_t[\mathcal{L}_R(t)]$, the gradient can be decomposed over time $\frac{\partial E}{\partial W} = \sum_t \frac{1}{T} \frac{\partial \mathcal{L}_R(t)}{\partial W}$, respectively. RTRL is an \textit{online algorithm} because it computes each term $\frac{\partial \mathcal{L}_R(t)}{\partial W}$ using dynamic updates of a tensor $\gamma$. This tensor measures the sensitivity of the $n$th node in the network on changes in each of the $n^2$ connection between nodes. Hence, it has a size of $n^3$. The update equation of $\gamma$ is deterministic and closed form. However, the specific form depends on the equation of the recurrence.  

In Appendix C, we derive the update equation of $\gamma$ for the recurrent gated circuit (RGC) and observe that its memory storage and computational cost is only $\mathcal{O}(n^2)$ due to the specific structure of the recurrence. More precisely, we find 

\begin{equation}
    \frac{\partial \mathcal{L}_R(t)}{\partial W} = \frac{\partial \mathcal{L}_R(t)}{\partial s(t-1)}^T \odot \Gamma(t-1).
\label{integrated-gradient}
\end{equation}

where $\Gamma \in \mathbb{R}^{n \times n}$ is a sensitivity matrix and $s \in \mathbb{R}^{n}$ is the current state of the network. The forward recursion equation of $\Gamma$ is presented in Appendix C (see Eq. \ref{eq:temporal_grad}). 

In Appendix D, we show that the forward recursion for $\Gamma$ provides an exact gradient computation for a more general class of recurrent networks which satisfy a \textit{two-points interactions property}. This stands in contrast to previous versions of forward propagation methods that did not appreciate this conditions and therefore obtain only approximate forward gradient computations \cite{marschall_unified_2019,bellec_solution_2020,irie_exploring_2024,gerstner_eligibility_2018}.

Note that $\mathcal{L}_R(t)$ is the instantaneous loss. Equation (\ref{integrated-gradient}) indicates that one obtains the gradient for learning by combining the sensitivity tensor with the dependence of the instantaneous loss on the current network state. If the loss only becomes available at a latter time, say at the end of a sequence, the sensitivity vector integrates the effects of the weights on the future network states. Therefore, there is no need to propagate the error back in time. In Appendix E we show that this approach works, provided the total loss is a sum (of functions) of the instantaneous losses. For instance, the instantaneous losses can be zero until the end of a sequence. If instantaneous losses are available, then gradient updates can be applied online at every time step (as in stochastic gradient descent), or after batches of data have been observed, or at the end of a sequence, with the usual relative merits of each approach.  

Finally, note that factor $\frac{\partial \mathcal{L}_R(t)}{\partial s(t-1)}$ can be straightforwardly computed because there is no recurrent computation between $s(t-1)$ and $\mathcal{L}_R(t)$. However, if the network is a multilayer hierarchy, this term does require standard (spatial) backpropagation.

\subsection{Empirical demonstration of real-time learning in R-JEPA}

\begin{figure}[ht]
\centering
\includegraphics[scale=0.45]{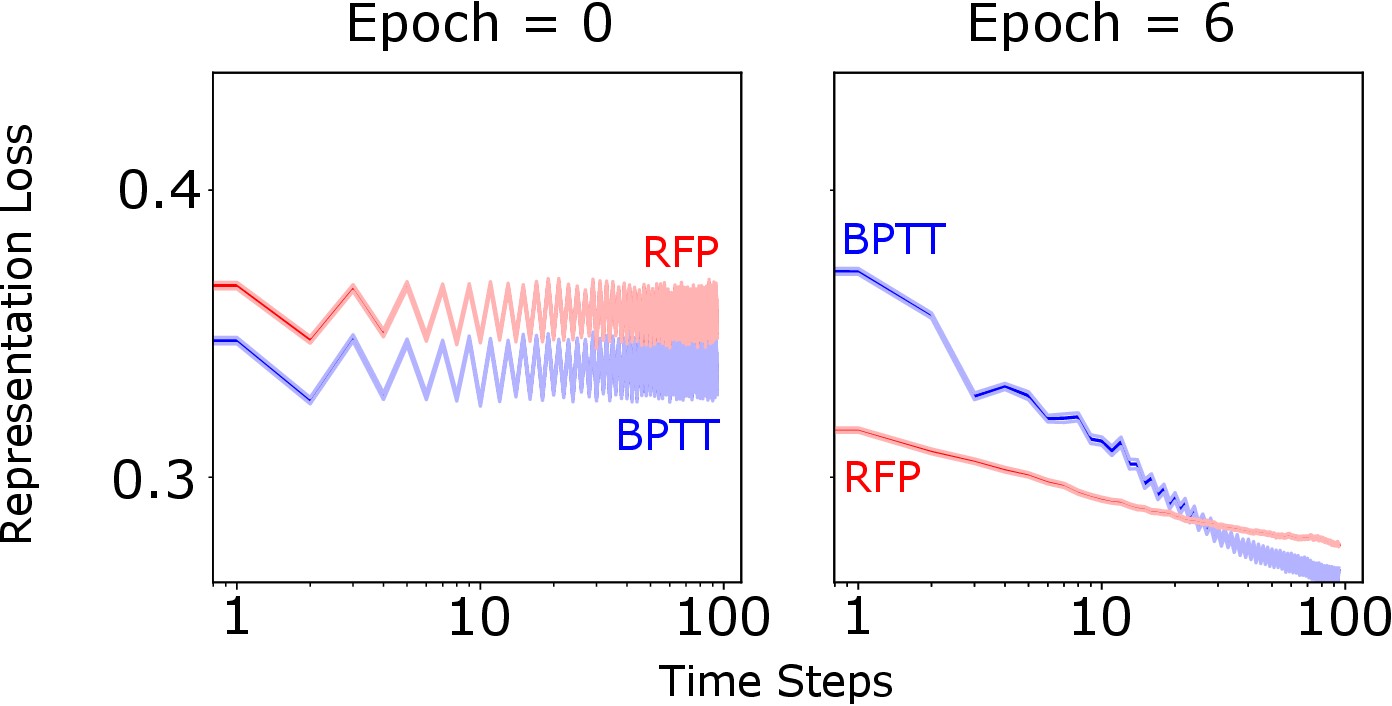}
\caption{\textbf{Prediction error as a function of time in video}.  Representation Loss indicates the ability of the recurrent network to predict the content of the next fixation (image patch). Curves are the average over 7000 fixation sequences in the test data. Network behavior is show before (Epoch=0) and after learning (Epoch=6). Time steps indicates the number of images patches (fixations) from the start of the recurrent iteration, i.e. the start of the test fixation sequences. The drop with time steps indicates that the network accumulates information in the context vector allowing it progressively improve its prediction}.
\label{fig:loss}
\end{figure}

We now empirically evaluate and compare the training of R-JEPA using both algorithms: BPTT and RFP. We used the same data and network structure as in the empirical evaluation above. The network is trained on sequences of length 100 of consecutive fixations selected at random from the full-length films. We had in total 29000 segments for training and 7000 for testing. Before training the network shows no evidence of integration of information across time, i.e. fixations (Fig.~\ref{fig:loss}). After learning, the prediction error decreases with time (the sequence of fixations), suggesting that the information carried by the context vector allows the network to progressively improve its prediction of what it will "see" next in a movie scene. 

We use convolutional layers from a pretrained ResNet-50, which are frozen during training. The weights of the predictor (MLP) were randomly initialized under the condition that $\hat{h}(t+1) \approx h(t)$ (i.e., equivalent to SimCLR). The weights of the RGC were initialized to zero, ensuring no temporal dynamics at Epoch 0 (see Fig. \ref{fig:loss} left). When the network is trained with BPTT, $\mathcal{L}_R(t)$ tends to increase at the first time step ($t=1$) before decreasing. This is expected behavior for BPTT, as at the start of the sequence, the encoder lacks sufficient feedback from later time steps to make accurate predictions. As the sequence progresses, the feedback effects accumulate, allowing the model to improve. In the case of RFP, the uniform decrease in the $\mathcal{L}(t)$ is a common characteristic (see Fig. \ref{fig:loss} right).

\section{Conclusion}
We have introduced a recurrent version of JEPA motivated by recurrence observed in biological vision networks. We made two important theoretical contributions. First, we demonstrate that next-step prediction with a stop-gradient achieves diverse representation thus avoiding representational collapse. Second, we derived general conditions for recurrent networks to be efficiently trained online with Recurrent Forward Propagation. Future work will explore more complete recurrent architectures with feedback at all areas of the visual processing hierarchy (as indicated in Fig.~\ref{fig:summary}), and will evaluate the performance of trained networks on downstream video processing tasks. To facilitate this, we are sharing the ongoing development of the R-JEPA architecture at \hyperlink{https://github.com/OsvaVelarde}{Github}. We also intend to compare network representations with neural activity similar to previous efforts \cite{yamins_using_2016,nayebi_recurrent_2022,zhuang_unsupervised_2021}, as soon as neural data during free-viewing of videos becomes available. 

\section{Acknowledgment}
We would like to thanks Jens Madsen for providing saccade data from participants watching movies and the corresponding time-aligned video files used in the empirical demonstrations here. 

\appendix

\appendixname{Appendix}

\section{Proof of Theorem 1}

Consider the following R-JEPA

\begin{align*} 
c(t) &=  C(x(t), c(t-1); \theta^{(C)}) & \text{R-Enconder: RNN} \\
h(t) &=  c^{High}(t) & \text{R-Enconder: Embedding} \\
\hat{a}(t) &=  W_A c^{Low}(t) & \text{Action Predictor} \\
\hat{h}(t+1) &=  W_{Gh} h(t) + W_{Ga} \hat{a}(t) & \text{Representation Predictor} \\
E &= \frac{1}{2} \mathbb{E}_t[\lambda_1 ||h(t) - \hat{h}(t)||^2 + \lambda_2 ||a(t) - \hat{a}(t)||^2] & \text{Loss Function}
\end{align*}
where embedding $F$, action predictor $A$, and representation predictor $G$ are linear maps. Here, $c(t) = [c^{Low}(t), c^{High}(t)]$ with $c^{Low} \in \mathbb{R}^{d_{cL}}$ and $c^{High} \in \mathbb{R}^{d_{cH}}$, and $x(t) \in \mathbb{R}^{d_0}$, $\hat{a}(t) \in \mathbb{R}^{d_A}$.

\subsection{Variation of loss function respect to weights}
Let's calculate the derivatives $\frac{\partial E}{\partial W_{Gh}}$ and $\frac{\partial E}{\partial W_{Ga}}$. First, note that

\begin{align*}
    \mathbb{E}_t[||h(t) - \hat{h}(t)||^2] &= \mathbb{E}_t[\text{Tr}[(h(t) - \hat{h}(t)) (h(t) - \hat{h}(t))^T]]\\
    &= \mathbb{E}_t[\text{Tr}[h(t)h(t)^T - 2  \hat{h}(t)h(t)^T + \hat{h}(t) \hat{h}(t)^T]]\\
    &= \text{Tr}[\mathbb{E}_t[h(t)h(t)^T - 2  \hat{h}(t)h(t)^T + \hat{h}(t) \hat{h}(t)^T]] \\
    &= \text{Tr}[R_0 - 2  W_{Gh} R_1 + W_{Gh} R_0 W_{Gh}^T + W_{Ga} W_A R_{clow} W_A^T W_{Ga}^T\\
    & - 2 W_{Ga} W_A \mathbb{E}_t[c^{Low}(t-1) h(t)^T] \\
    & + 2 W_{Gh} \mathbb{E}_t[h(t) c^{Low}(t)^T] W_A^T W_{Ga}^T]
\end{align*}
where $R_0 = \mathbb{E}_t[h(t)h^T(t)]$ and $R_1 = \mathbb{E}_t[h(t)h^T(t-1)]$.

Therefore,
\begin{align*}
    \frac{\partial E}{\partial W_{Gh}}  &= \frac{\lambda_1}{2} \frac{\partial}{\partial W_{Gh}} \text{Tr}[- 2 W_{Gh} R_1 + W_{Gh} R_0 W_{Gh}^T + 2 W_{Gh} \mathbb{E}_t[h(t) c^{Low}(t)^T] W_A^T W_{Ga}^T] \\
    &= \lambda_1 [- R_1 + W_{Gh} R_0 + W_{Ga} W_A \mathbb{E}_t[c^{Low}(t)h(t)^T]]
\end{align*}
and

\begin{align*}
    \frac{\partial E}{\partial W_{Ga}}  &= \frac{\lambda_1}{2} \frac{\partial}{\partial W_{Ga}} \text{Tr}[W_{Ga} W_A R_{cH} W_A^T W_{Ga}^T - 2 W_{Ga} W_A \mathbb{E}_t[c^{Low}(t-1) h(t)^T] \\
    & + 2 W_{Ga} W_A \mathbb{E}_t[c^{Low}(t) h(t)^T] W_{Gh}^T] \\
    & = \lambda_1 [W_{Ga} W_A R_{clow} W_A^T - \mathbb{E}_t[h(t) c^{Low}(t-1)^T] W_A^T + W_{Gh} \mathbb{E}_t[h(t) c^{Low}(t)^T] W_A^T]. 
\end{align*}

\subsection{Variation of loss function respect to representations}

To calculate the derivative $\frac{\partial E}{\partial h(q)}$ based on the following equality,

\begin{align*}
    \mathbb{E}_t[||h(t) - \hat{h}(t)||^2] &= \frac{1}{T} \sum_{t=1}^{T} ||\text{stop}[h(t)] - \hat{h}(t)||^2 \\
    &= \frac{1}{T} \sum_{t=1}^T ||\text{stop}[h(t)] -  W_{Gh} h(t-1) - W_{Ga} W_A c^{Low}(t-1)||^2.
\end{align*}

The result is
\begin{align*}
    \frac{\partial E}{\partial h(q)}  &= \frac{1}{T} \frac{\partial}{\partial h(q)} \sum_{t=1}^T ||\text{stop}[h(t)] -  W_{Gh} h(t-1) - W_{Ga} W_A c^{Low}(t-1)||^2 \\
    &= - \frac{1}{T} \sum_{t=1}^T [h(t) -  W_{Gh} h(t-1) - W_{Ga} W_A c^{Low}(t-1)]^T W_{Gh}^T \delta_{q,t-1}\\
    &= \frac{\Theta(0 \leq q \leq T-1)}{T} [-h(q+1) +  W_{Gh} h(q) + W_{Ga} W_A c^{Low}(q)]^T W_{Gh} \\
    &= \frac{\Theta(0 \leq q \leq T-1)}{T} [-h(q+1)^T+ h(q)^T W_{Gh}^T +  c^{Low}(q)^T W_A^T W_{Ga}^T]W_{Gh}.
\end{align*}

\subsection{Learning dynamics of R-JEPA}
In R-JEPA, the learning dynamics of the weights are obtained by minimizing $E$,

\begin{align*}
\delta h(t) &= \frac{\partial F}{\partial c(t)} \frac{\partial C}{\partial \theta_c} \delta \theta_c -\eta \frac{\partial F}{\partial c(q)} \frac{\partial C}{\partial \theta_c} \theta_c \\
\delta W_{Gh} &= -\frac{\partial E}{\partial W_{Gh}} - \eta  W_{Gh} \\
\delta W_{Ga} &= -\frac{\partial E}{\partial W_{Ga}} - \eta  W_{Ga}
\end{align*}

where $\eta$ is a weight decay parameter. This differential system is a good approximation of the limit of large batch sizes and small discrete time learning rates \cite{tian_understanding_2021}.

By the gradient descent, the term $\delta \theta_c = - \left(\frac{\partial E}{\partial \theta_c}\right)^T$ is equal to $-\left(\frac{\partial E}{\partial h(t)} \frac{\partial F}{\partial c(t)} \frac{\partial C}{\partial \theta_c}\right)^T$. Then, 

\begin{align}
    \delta h(t) &= - M \frac{\partial C}{\partial \theta_c} \frac{\partial C}{\partial \theta_c}^T M^T \frac{\partial E}{\partial h(t)}^T -\eta M \frac{\partial C}{\partial \theta_c} \theta_c \\
    \delta W_{Gh} &= -\frac{\partial E}{\partial W_{Gh}} - \eta  W_{Gh} \\
    \delta W_{Ga} &= -\frac{\partial E}{\partial W_{Ga}} - \eta  W_{Ga}
\end{align}

where $M = \frac{\partial F}{\partial c(t)} = [\mathbf{0}|\mathbf{I}] \in \mathbb{R}^{d_{cH} \times d_c}$ where $d_c = d_{cL} + d_{cH}$. We denote $H = \frac{1}{\sqrt{T}}[h(1),...,h(T)]$; then,

\begin{align*}
    \delta H H^T &= \mathbb{E}_t(\delta h(t) h^T(t)) \\
    &= \mathbb{E}_t (-M \frac{\partial C}{\partial \theta_c} \frac{\partial C}{\partial \theta_c}^T M^T [ W_{Gh}^T W_{Gh} h(t) h(t)^T \\
    & +  W_{Gh}^T W_{Ga} W_A c^{Low}(t) h(t)^T - W_{Gh}^T h(t+1) h(t)^T] \\
    & -\eta M \frac{\partial C}{\partial \theta_c} \theta_c h(t)^T).
\end{align*}

\subsection{Effect of recurrent encoder}
To analyze the learning dynamics, we have to calculate the term $\frac{\partial C}{\partial \theta_c}$. This derivative depends of the expression of $C$. For simplicity, we use Time Decay Units $c_{l,t} = \tau_l c_{l,t-1} + P^{(l)} c_{l-1,t}$ with $c_l \in \mathbb{R}^{d_l}$ and $c_0 = x$, $l=1,...,N+1$. Note that $c^{High}=c_{N+1}$ and $c^{Low}=[c_1,...,c_N]$.

We use first-order approximation of BPTT to calculate $\frac{\partial C}{\partial \theta_c}$ where 

$\theta_c = [\tau_1, \tau_2,..., \tau_{N+1}, P^{(1)}_{11}, ..., P^{(1)}_{1d_0}, P^{(1)}_{21}, ..., P^{(1)}_{2d_0}, ..., P^{(1)}_{d_1 1}, ..., P^{(1)}_{d_1d_0} ...]$.

We obtain that
\begin{equation*}
\frac{\partial C}{\partial \theta_c} \approx \begin{bmatrix}
  c_{1,t-1}e_1^T & 0 & ... & 0 & I_{d1} \otimes x^T_t & 0 & ... & 0 \\
  0 & c_{2,t-1}e_2^T & ... & 0 & 0 & I_{d2} \otimes c_{1,t}^T & ... & 0 \\
  ... & ... & ... & 0 & ... & ... & ... & ... \\
  0 & 0 & ... & c_{N+1,t-1}e_{N+1}^T & 0 & 0 & ... & I_{d_{N+1}} \otimes c_{N,t}^T
\end{bmatrix}.
\end{equation*}

Therefore,
\begin{align*}
L(t) = \frac{\partial C}{\partial \theta_c}\frac{\partial C}{\partial \theta_c}^T \approx \begin{bmatrix}
  c_{1,t-1}c_{1,t-1}^T & 0 & ... & 0  \\
  0 & c_{2,t-1}c_{2,t-1}^T & ... & 0 \\
  ... & ... & ... & 0 \\
  0 & 0 & ... & c_{N+1,t-1}c_{N+1,t-1}^T 
\end{bmatrix} \\
+ \begin{bmatrix}
  I_{d1} ||x_t||^2 & 0 & ... & 0  \\
  0 & I_{d2} ||c_{1,t}||^2 & ... & 0 \\
  ... & ... & ... & 0 \\
  0 & 0 & ... & I_{d_{N+1}} ||c_{N,t}||^2 
\end{bmatrix},
\end{align*}
and
\begin{equation*}
\frac{\partial C}{\partial \theta_c}\theta_c \approx c(t).
\end{equation*}

Also, $M L(t) M^T = h(t-1)h(t-1)^T + ||c_{N,t}||^2 \mathbf{I}$. We will use the normalization $||c_N(t)||^2=1$.

\subsection{Balancing}
Let's calculate the following variations

\begin{align*}
    \delta H H^T &= \mathbb{E}_t(\delta h(t) h^T(t)) \\
    &= \mathbb{E}_t ( - [h(t-1)h(t-1)^T + \mathbf{I}] [W_{Gh}^T W_{Gh} h(t) h(t)^T \\
    & +  W_{Gh}^T W_{Ga} W_A c^{Low}(t) h(t)^T - W_{Gh}^T h(t+1) h(t)^T] \\
    & -\eta h(t) h(t)^T ) \\
    & = - W_{Gh}^T W_{Gh} R_0 + W_{Gh}^T R_1 -\eta R_0 - W_{Gh}^T W_{Ga} W_A \mathbb{E}_t(c^{Low}(t) h(t)^T) + Y,
\end{align*}

\begin{align*}
    W_{Gh}^T \delta W_{Gh} &= -W_{Gh}^T\frac{\partial E}{\partial W_{Gh}} - \eta  W_{Gh}^T W_{Gh}\\
    &= \lambda_1 [W_{Gh}^T R_1 - W_{Gh}^T W_{Gh} R_0 - W_{Gh}^T W_{Ga} W_A \mathbb{E}_t[c^{Low}(t)h(t)^T]] - \eta  W_{Gh}^T W_{Gh} \\
    &= \lambda_1 [\delta H H^T + \eta R_0 - Y] - \eta  W_{Gh}^T W_{Gh},
\end{align*}

and 

\begin{align*}
    W_{Ga}^T \delta W_{Ga} &= -W_{Ga}^T\frac{\partial E}{\partial W_{Ga}} - \eta  W_{Ga}^T W_{Ga}\\
    &= - \lambda_1 W_{Ga}^T  [W_{Ga} W_A R_{clow} W_A^T - \mathbb{E}_t[h(t) c^{Low}(t-1)^T] W_A^T \\
    & + W_{Gh} \mathbb{E}_t[h(t) c^{Low}(t)^T] W_A^T] - \eta  W_{Ga}^T W_{Ga}.
\end{align*}

Finally,
\begin{equation*}
W_{Gh}^T \delta W_{Gh} + W_{Gh} \delta W_{Gh}^T + 2\eta  W_{Gh}^T W_{Gh} = \lambda_1 [\delta H H^T + \delta H^T H + 2\eta R_0] - \lambda_1 [Y + Y^T].
\end{equation*}

Integrating the equation, we obtain $\delta (e^{2\eta s} W_{Gh}^T W_{Gh}) = \lambda_1 \delta (e^{2\eta s} H^T H) - \lambda_1 [Y + Y^T]$. In Appendix B, we show that $Y \propto \tau^2$. Under the stable regimen, $Y << 1$; then, 

\begin{equation}
W_{Gh}^T W_{Gh} = \lambda_1 H^T H, \qquad (s \rightarrow \infty)
\end{equation}

\section{Feature correlations}

Let the linear equation be,
\begin{equation}   
    \vec{c}(t) = U \vec{c}(t-1) + b(t)
\end{equation}
where the random variable $b(t)$ verifies 
\begin{enumerate}
    \item $T^{B}_{i,j,m,n}=\mathbb{E}[b_i(t)b_j(t)b_m(t)b_n(t)] = \Sigma_{ij} \Sigma_{mn} + \Sigma_{im} \Sigma_{jn} + \Sigma_{in} \Sigma_{jm} = f(i,j,m,n,\Sigma)$
    \item $T_{ijmn}(\Delta_1,\Delta_2,\Delta_3) = \mathbb{E}[c_i(t)c_j(t-\Delta_1)c_m(t-\Delta_2)c_n(t-\Delta_3)]$
\end{enumerate}
and $U \propto \text{diag}(\tau_i)$ with $\tau<1$. 

Then, we find 
\begin{align*}
    T_{ijmn}(0,0,0) &= \mathbb{E}[c_i(t)c_j(t)c_m(t)c_n(t)] \\
    & = \sum_{k_1,k_2,k_3,k_4} U_{ik_1}U_{jk_2}U_{mk_3}U_{nk_4} \mathbb{E}[c_{k_1}(t-1)c_{k_2}(t-1)c_{k_3}(t-1)c_{k_4}(t-1)] \\
    & + \mathbb{E}[b_i(t)b_j(t)b_m(t)b_n(t)] \\
    & = \sum_{k_1,k_2,k_3,k_4} U_{ik_1}U_{jk_2}U_{mk_3}U_{nk_4} T_{k_1,k_2,k_3,k_4}(0,0,0) + T^B_{ijmn}
\end{align*}

\begin{align*}
    T_{ijmn}(0,1,1) &= \mathbb{E}[c_i(t)c_j(t)c_m(t-1)c_n(t-1)] \\
    & = \sum_{k_1,k_2} U_{ik_1}U_{jk_2} \mathbb{E}[c_{k_1}(t-1)c_{k_2}(t-1)c_m(t-1)c_n(t-1)] \\
    & = \sum_{k_1,k_2} U_{ik_1}U_{jk_2} T_{k_1k_2mn}(0,0,0)
\end{align*}

\begin{align*}
    T_{ijmn}(1,2,2) &= \mathbb{E}[c_i(t)c_j(t-1)c_m(t-2)c_n(t-2)] \\
    & = \sum_{k_1} U_{ik_1} \mathbb{E}[c_{k_1}(t-1)c_j(t-1)c_m(t-2)c_n(t-2)] \\
    & = \sum_{k_1} U_{ik_1} T_{k_1jmn}(0,1,1) \\
    & = \sum_{k_1} U_{ik_1} [\sum_{k_2,k_3} U_{k_1k_2}U_{jk_3} T_{k_2k_3mn}(0,0,0)] \\
    & = \sum_{k_1,k_2} U^2_{ik_1}U_{jk_2} T_{k_1k_2mn}(0,0,0)
\end{align*}

If we express the tensor $T$ as a vector $\textbf{vec}(T)$, we obtain

\begin{equation}
    \textbf{vec}(T)(0,0,0) = [I - U \otimes U \otimes U \otimes U]^{-1}\textbf{vec}(T^B)
\end{equation}

\begin{equation}
    \textbf{vec}(T)(0,1,1) = [U \otimes U \otimes I \otimes I] [I - U \otimes U \otimes U \otimes U]^{-1}\textbf{vec}(T^B)
\end{equation}

\begin{equation}
    \textbf{vec}(T)(1,2,2) = [U^2 \otimes U \otimes I \otimes I] [I - U \otimes U \otimes U \otimes U]^{-1}\textbf{vec}(T^B)
\end{equation}

Finally, 

\begin{align*}
    \textbf{vec}(T)(0,0,0) &= \textbf{vec}(T^B), \\
    \textbf{vec}(T)(0,1,1) &= \text{O}(\tau^2), \\
    \textbf{vec}(T)(1,2,2) &= \text{O}(\tau^3).
\end{align*}

\section{Real-Time Recurrent Learning for Recurrent Gated Cells}

Based on the chain rule, we know $\frac{\partial \mathcal{L}_R(t)}{\partial W^k_{pq}} = \frac{\partial \mathcal{L}_R(t)}{\partial s(t-1)} \frac{\partial s(t-1)}{\partial W^k_{pq}}$. Now, the point is to calculate $\frac{\partial s(t-1)}{\partial W^k_{pq}}$.

First, note that a compact way to write the recurrence equation of RGC is
\begin{equation}
    c^{(\nu)}_{i}(t) = (1-a^{(\nu)}_{i}) x_i(t) + b^{(\nu)}_{i} c^{(\nu)}_{i}(t-1),    \qquad i=1,...,n
\end{equation}
where $c^{(0)}(t)=s(t)$, $c^{(1)}(t)=m(t)$, and

\begin{align*}
    a^{(\nu)}_{i} &= f(\sum_j W_{ij}^{(2\nu+1)} c^{(1-\nu)}_{j}(t-1)),\\
    b^{(\nu)}_{i} &= f(\sum_j W_{ij}^{(2\nu)} c^{(\nu)}_{j}(t-1)).
\end{align*}

where we use the index $k=0,1,2,3$ to indicate the pairs $ss$, $ms$, $mm$, and $sm$, respectively. 

Let define $\gamma^{\nu,k}_{ipq}(t) = \frac{\partial c_{i,\nu}(t)}{\partial W^{k}_{pq}}$, then:
\begin{align*}
    \gamma^{\nu,k}_{ipq}(t) &= - x_i(t) [1-a^{(\nu)2}_{i}] [\delta_{ip} \delta_{k,2\nu+1} c^{(1-\nu)}_{q}(t-1) + \sum_j W_{ij}^{(2\nu+1)} \gamma^{1-\nu,k}_{jpq} (t-1)] \\
    & + b^{(\nu)}_{i}  \gamma^{\nu,k}_{ipq}(t-1) + [1-b_{i,\nu}^2]c^{(\nu)}_{i}(t-1)[\delta_{k,2\nu}\delta_{ip}c^{(\nu)}_{q}(t-1)  + \sum_j W_{ij}^{(2\nu)} \gamma^{\nu,k}_{jpq}(t-1)] \\
    & = \delta_{ip} [\delta_{k,2\nu} [1-b^{(\nu)2}_{i}] c^{(\nu)}_{i}(t-1) c^{(\nu)}_{q}(t-1) - \delta_{k,2\nu+1} [1-a^{(\nu)2}_{i}]  x_i(t) c^{(1-\nu)}_{q}(t-1)] \\
    & - [1-a^{(\nu)2}_{i}] x_i(t) \sum_j W_{ij}^{(2\nu+1)} \gamma^{1-\nu,k}_{jpq} (t-1) \\
    & + [1-b^{(\nu)2}_{i}] c^{(\nu)}_{i}(t-1)\sum_j W_{ij}^{(2\nu)} \gamma^{\nu,k}_{jpq}(t-1) \\
    & + b^{(\nu)}_{i}  \gamma^{\nu,k}_{ipq}(t-1).
\end{align*}

The initial condition $\gamma^{\nu,k}_{ipq}(0)=0$ implies $\gamma^{\nu,k}_{ipq}(t) = \Gamma^{\nu,k}_{pq}(t) \delta_{ip}$. Then, the the temporal evolution of $\gamma$ can be written as:
\begin{equation}
\begin{split}
    \Gamma^{\nu,k}(t) &= \mu^{\nu,0}(t) \odot \Gamma^{\nu,k}(t-1) + \mu^{\nu,1}(t) \odot \Gamma^{1-\nu,k}(t-1) + \delta_{k//2,\nu} J^{\nu,k\%2} \\
    \mu^{\nu,0}(t) &= [1 - b^{(\nu)2}] \odot c^{(\nu)}(t-1) \odot \text{diag}(W^{(2\nu)}) + b^{(\nu)} \\
    \mu^{\nu,1}(t) &= -[1 - a^{(\nu)2}] \odot I(t) \odot \text{diag}(W^{(2\nu+1)}) \\
    J^{\nu,0}(t) &= [1-b^{(\nu)2}] \odot c^{(\nu)}(t-1) c^{(\nu)T}(t-1) \\
    J^{\nu,1}(t) &= -[1-a^{(\nu)2}] \odot I(t) c^{(1-\nu)T}(t-1) 
\end{split}
\label{eq:temporal_grad}
\end{equation}

Finally, 
\begin{align*}
    \frac{\partial \mathcal{L}_R(t)}{\partial W^k_{pq}} &= \frac{\partial \mathcal{L}_R(t)}{\partial s(t-1)} \frac{\partial s(t-1)}{\partial W^k_{pq}}\\
    & = \sum_i \frac{\partial \mathcal{L}_R(t)}{\partial s_i(t-1)} \gamma^{0,k}_{ipq}(t-1) \\
    &= \frac{\partial \mathcal{L}_R(t)}{\partial s_p(t-1)} \Gamma^{0,k}_{pq}(t-1).
\end{align*}

\section{Two-point Interaction Property}

Here we show that networks with two-point interactions RTRL reduces in complexity. Given a recurrent network with $n$ units (i.e., context vector $c \in \mathbb{R}^n$) and $P$ parameters (parameter vector $w \in \mathbb{R}^P$), we define $\gamma(t) = \frac{\partial c(t)}{\partial w} \in \mathbb{R}^{n \times P}$, $J(t) = \frac{\partial c(t)}{\partial c(t-1)} \in \mathbb{R}^{n \times n}$, and $R(t) = \frac{\partial c(t)}{\partial w^{(t)}} \in \mathbb{R}^{n \times P}$.

The recursion for $\gamma$ is:
\begin{equation}
    \gamma(t) = J(t) \gamma(t-1) + R(t)
\label{eq:gamma}
\end{equation}

Suppose each parameter $w_p$ represents the effect of a unit $j$ on unit $i$. We will say that this parameter satisfies the property of \textit{two-point interaction}, if no other unit $\hat{j}$ influences $i$ through $w_p$, and no other unit $\hat{i}$ receives information from $j$ via $w_p$. It means $p = (i, j)$, and implies
\begin{equation}
    R_{k,p}(t) = R_{k,(i,j)}(t) = \frac{\partial c_k(t)}{\partial w_{i,j}^{(t)}} = \delta_{ki} \hat{R}_{ij}(t).
\end{equation}

Then, Eq. \ref{eq:gamma} will be:
\begin{equation}
    \gamma_{k,(ij)}(t) = \sum_m J_{k,m}(t) \gamma_{m,(ij)}(t-1) + \delta_{ki} \hat{R}_{ij}(t)
\label{eq:gamma2}
\end{equation}

Using the initial condition $\gamma(0)=0$, we obtain $\gamma_{k,(ij)}(t) = \delta_{ki} \Gamma_{ij}(t)$ and

\begin{equation}
    \Gamma_{ij}(t) = J_{ii}(t) \Gamma_{ij}(t-1) + \hat{R}_{ij}(t)
\end{equation}

Under the two-point interaction hypothesis, the size of $\Gamma$ and the number of necessary operations is $n^2$ for this case. This is a complexity reduction of the general case of RTRL (i.e. $np \sim n^3$). We refer to this more efficient learning rule as Recurrent Forward Propagation. 

\section{About Loss function E}
Typically, in an RNN, the loss function \( E \) is a combination of instantaneous errors \( \mathcal{L}(t) \). For example, in R-JEPA, \( E = \frac{1}{T} \sum^T_{t=1} \mathcal{L}(t) \). In a biological model, it is expected that the instantaneous errors are independent of each other, which can be modeled with
\begin{equation}
    E = \sum^T_{t=1} \psi(t,\mathcal{L}(t)).
\end{equation}

These types of loss functions ensure that RTRL can be applied, as each term in the following sum can be computed online,
\begin{equation}
    \frac{\partial E}{\partial W} = \sum^T_{t=1} \frac{\partial \psi}{\partial \mathcal{L}} \left.\right|_{t,\mathcal{L}(t)} \frac{\partial \mathcal{L}(t)}{\partial W}
\end{equation}

%Generalized Linear Gating

%\bibliography{references}

\begin{thebibliography}{10}

\bibitem{garrido_learning_2024}
Quentin Garrido, Mahmoud Assran, Nicolas Ballas, Adrien Bardes, Laurent Najman, and Yann LeCun.
\newblock Learning and {Leveraging} {World} {Models} in {Visual} {Representation} {Learning}, March 2024.
\newblock arXiv:2403.00504.

\bibitem{nayebi_recurrent_2022}
Aran Nayebi, Javier Sagastuy-Brena, Daniel~M. Bear, Kohitij Kar, Jonas Kubilius, Surya Ganguli, David Sussillo, James~J. DiCarlo, and Daniel L.~K. Yamins.
\newblock Recurrent {Connections} in the {Primate} {Ventral} {Visual} {Stream} {Mediate} a {Trade}-{Off} {Between} {Task} {Performance} and {Network} {Size} {During} {Core} {Object} {Recognition}.
\newblock {\em Neural Computation}, 34(8):1652--1675, July 2022.

\bibitem{werbos_backpropagation_1990}
P.J. Werbos.
\newblock Backpropagation through time: what it does and how to do it.
\newblock {\em Proceedings of the IEEE}, 78(10):1550--1560, October 1990.
\newblock Conference Name: Proceedings of the IEEE.

\bibitem{williams_learning_1989}
Ronald~J. Williams and David Zipser.
\newblock A {Learning} {Algorithm} for {Continually} {Running} {Fully} {Recurrent} {Neural} {Networks}.
\newblock {\em Neural Computation}, 1(2):270--280, June 1989.
\newblock Conference Name: Neural Computation.

\bibitem{kong_transsaccadic_2021}
Garry Kong, Lisa~M. Kroell, Sebastian Schneegans, David Aagten-Murphy, and Paul~M. Bays.
\newblock Transsaccadic integration relies on a limited memory resource.
\newblock {\em Journal of Vision}, 21(5):24, May 2021.

\bibitem{oostwoud_wijdenes_evidence_2015}
Leonie Oostwoud~Wijdenes, Louise Marshall, and Paul~M. Bays.
\newblock Evidence for {Optimal} {Integration} of {Visual} {Feature} {Representations} across {Saccades}.
\newblock {\em The Journal of Neuroscience: The Official Journal of the Society for Neuroscience}, 35(28):10146--10153, July 2015.

\bibitem{vision_information_1990}
National Research Council (US) Committee~on Vision.
\newblock Information {Processing} in the {Primate} {Visual} {System}.
\newblock In {\em Advances in the {Modularity} of {Vision}: {Selections} {From} a {Symposium} on {Frontiers} of {Visual} {Science}}. National Academies Press (US), 1990.

\bibitem{celeghin_convolutional_2023}
Alessia Celeghin, Alessio Borriero, Davide Orsenigo, Matteo Diano, Carlos~Andrés Méndez~Guerrero, Alan Perotti, Giovanni Petri, and Marco Tamietto.
\newblock Convolutional neural networks for vision neuroscience: significance, developments, and outstanding issues.
\newblock {\em Frontiers in Computational Neuroscience}, 17, July 2023.
\newblock Publisher: Frontiers.

\bibitem{schmidt_recurrent_2019}
Robin~M. Schmidt.
\newblock Recurrent {Neural} {Networks} ({RNNs}): {A} gentle {Introduction} and {Overview}, November 2019.
\newblock arXiv:1912.05911.

\bibitem{gilbert_top-down_2013}
Charles~D. Gilbert and Wu~Li.
\newblock Top-down influences on visual processing.
\newblock {\em Nature Reviews. Neuroscience}, 14(5):350--363, May 2013.

\bibitem{bai_empirical_2018}
Shaojie Bai, J.~Zico Kolter, and Vladlen Koltun.
\newblock An {Empirical} {Evaluation} of {Generic} {Convolutional} and {Recurrent} {Networks} for {Sequence} {Modeling}, April 2018.
\newblock arXiv:1803.01271.

\bibitem{alom_inception_2017}
Md~Zahangir Alom, Mahmudul Hasan, Chris Yakopcic, and Tarek~M. Taha.
\newblock Inception {Recurrent} {Convolutional} {Neural} {Network} for {Object} {Recognition}, April 2017.
\newblock arXiv:1704.07709.

\bibitem{donahue_long-term_2016}
Jeff Donahue, Lisa~Anne Hendricks, Marcus Rohrbach, Subhashini Venugopalan, Sergio Guadarrama, Kate Saenko, and Trevor Darrell.
\newblock Long-term {Recurrent} {Convolutional} {Networks} for {Visual} {Recognition} and {Description}, May 2016.
\newblock arXiv:1411.4389.

\bibitem{assran_self-supervised_2023}
Mahmoud Assran, Quentin Duval, Ishan Misra, Piotr Bojanowski, Pascal Vincent, Michael Rabbat, Yann LeCun, and Nicolas Ballas.
\newblock Self-{Supervised} {Learning} from {Images} with a {Joint}-{Embedding} {Predictive} {Architecture}, April 2023.
\newblock arXiv:2301.08243.

\bibitem{dawid_introduction_2024}
Anna Dawid and Yann LeCun.
\newblock Introduction to latent variable energy-based models: a path toward autonomous machine intelligence.
\newblock {\em Journal of Statistical Mechanics: Theory and Experiment}, 2024(10):104011, October 2024.
\newblock Publisher: IOP Publishing.

\bibitem{chen_simple_2020}
Ting Chen, Simon Kornblith, Mohammad Norouzi, and Geoffrey Hinton.
\newblock A {Simple} {Framework} for {Contrastive} {Learning} of {Visual} {Representations}, July 2020.
\newblock arXiv:2002.05709.

\bibitem{chen_exploring_2020}
Xinlei Chen and Kaiming He.
\newblock Exploring {Simple} {Siamese} {Representation} {Learning}, November 2020.
\newblock arXiv:2011.10566.

\bibitem{zhuang_unsupervised_2021}
Chengxu Zhuang, Siming Yan, Aran Nayebi, Martin Schrimpf, Michael~C. Frank, James~J. DiCarlo, and Daniel L.~K. Yamins.
\newblock Unsupervised neural network models of the ventral visual stream.
\newblock {\em Proceedings of the National Academy of Sciences}, 118(3):e2014196118, January 2021.
\newblock Publisher: Proceedings of the National Academy of Sciences.

\bibitem{zbontar_barlow_2021}
Jure Zbontar, Li~Jing, Ishan Misra, Yann LeCun, and Stéphane Deny.
\newblock Barlow {Twins}: {Self}-{Supervised} {Learning} via {Redundancy} {Reduction}, June 2021.
\newblock arXiv:2103.03230.

\bibitem{bardes_vicreg_2022}
Adrien Bardes, Jean Ponce, and Yann LeCun.
\newblock {VICReg}: {Variance}-{Invariance}-{Covariance} {Regularization} for {Self}-{Supervised} {Learning}, January 2022.
\newblock arXiv:2105.04906.

\bibitem{marschall_unified_2019}
Owen Marschall, Kyunghyun Cho, and Cristina Savin.
\newblock A {Unified} {Framework} of {Online} {Learning} {Algorithms} for {Training} {Recurrent} {Neural} {Networks}, July 2019.
\newblock arXiv:1907.02649.

\bibitem{irie_exploring_2024}
Kazuki Irie, Anand Gopalakrishnan, and Jürgen Schmidhuber.
\newblock Exploring the {Promise} and {Limits} of {Real}-{Time} {Recurrent} {Learning}, February 2024.
\newblock arXiv:2305.19044.

\bibitem{gerstner_eligibility_2018}
Wulfram Gerstner, Marco Lehmann, Vasiliki Liakoni, Dane Corneil, and Johanni Brea.
\newblock Eligibility {Traces} and {Plasticity} on {Behavioral} {Time} {Scales}: {Experimental} {Support} of {NeoHebbian} {Three}-{Factor} {Learning} {Rules}.
\newblock {\em Frontiers in Neural Circuits}, 12:53, July 2018.

\bibitem{bellec_solution_2020}
Guillaume Bellec, Franz Scherr, Anand Subramoney, Elias Hajek, Darjan Salaj, Robert Legenstein, and Wolfgang Maass.
\newblock A solution to the learning dilemma for recurrent networks of spiking neurons.
\newblock {\em Nature Communications}, 11(1):3625, July 2020.
\newblock Publisher: Nature Publishing Group.

\bibitem{yamins_using_2016}
Daniel L.~K. Yamins and James~J. DiCarlo.
\newblock Using goal-driven deep learning models to understand sensory cortex.
\newblock {\em Nature Neuroscience}, 19(3):356--365, March 2016.
\newblock Publisher: Nature Publishing Group.

\bibitem{nayebi_task-driven_2018}
Aran Nayebi, Daniel Bear, Jonas Kubilius, Kohitij Kar, Surya Ganguli, David Sussillo, James~J. DiCarlo, and Daniel L.~K. Yamins.
\newblock Task-{Driven} {Convolutional} {Recurrent} {Models} of the {Visual} {System}, October 2018.
\newblock arXiv:1807.00053.

\bibitem{grill_bootstrap_2020}
Jean-Bastien Grill, Florian Strub, Florent Altché, Corentin Tallec, Pierre~H. Richemond, Elena Buchatskaya, Carl Doersch, Bernardo~Avila Pires, Zhaohan~Daniel Guo, Mohammad~Gheshlaghi Azar, Bilal Piot, Koray Kavukcuoglu, Rémi Munos, and Michal Valko.
\newblock Bootstrap your own latent: {A} new approach to self-supervised {Learning}, September 2020.
\newblock arXiv:2006.07733.

\bibitem{liu_bridging_2022}
Kang-Jun Liu, Masanori Suganuma, and Takayuki Okatani.
\newblock Bridging the {Gap} from {Asymmetry} {Tricks} to {Decorrelation} {Principles} in {Non}-contrastive {Self}-supervised {Learning}.
\newblock {\em Advances in Neural Information Processing Systems}, 35:19824--19835, December 2022.

\bibitem{zhang_how_2022}
Chaoning Zhang, Kang Zhang, Chenshuang Zhang, Trung~X. Pham, Chang~D. Yoo, and In~So Kweon.
\newblock How {Does} {SimSiam} {Avoid} {Collapse} {Without} {Negative} {Samples}? {A} {Unified} {Understanding} with {Self}-supervised {Contrastive} {Learning}, March 2022.
\newblock arXiv:2203.16262.

\bibitem{ellenberger_backpropagation_2024}
Benjamin Ellenberger, Paul Haider, Jakob Jordan, Kevin Max, Ismael Jaras, Laura Kriener, Federico Benitez, and Mihai~A. Petrovici.
\newblock Backpropagation through space, time, and the brain, July 2024.
\newblock arXiv:2403.16933.

\bibitem{rumelhart_learning_1986}
David~E. Rumelhart, Geoffrey~E. Hinton, and Ronald~J. Williams.
\newblock Learning representations by back-propagating errors.
\newblock {\em Nature}, 323(6088):533--536, October 1986.
\newblock Publisher: Nature Publishing Group.

\bibitem{lillicrap_backpropagation_2019}
Timothy~P Lillicrap and Adam Santoro.
\newblock Backpropagation through time and the brain.
\newblock {\em Current Opinion in Neurobiology}, 55:82--89, April 2019.

\bibitem{hebb_organization_2002}
D.~O. Hebb.
\newblock {\em The {Organization} of {Behavior}: {A} {Neuropsychological} {Theory}}.
\newblock Psychology Press, New York, April 2002.

\bibitem{tian_understanding_2021}
Yuandong Tian, Xinlei Chen, and Surya Ganguli.
\newblock Understanding self-supervised {Learning} {Dynamics} without {Contrastive} {Pairs}, October 2021.
\newblock arXiv:2102.06810.

\end{thebibliography}

\bibliographystyle{unsrt}

%%%%%%%%%%%%%%%%%%%%%%%%%%%%%%%%%%%%%%%%%%%%%%%%%%%%%%%%%%%%%%%%%%%%%%%%%%%%%%%%%%%%%%%%%%%%%%%%%%%%%%%%%%%%%%%%%%%%%%%%%%%%%%%%%
\end{document}